\documentclass[11pt,letterpaper]{article}
\usepackage{emnlp2016}
\usepackage{times}
\usepackage{latexsym}

\emnlpfinalcopy

\def\emnlppaperid{191}

\usepackage{array}
\def\thline{\noalign{\hrule height 1pt}}

\usepackage{bm}

\usepackage{amsmath}
\usepackage{amsfonts}

\usepackage{graphicx}

\newcommand{\figref}[1]{Figure \ref{#1}}
\newcommand{\tabref}[1]{Table \ref{#1}}

\newcommand{\secref}[1]{Section \ref{#1}}

\usepackage[tight]{subfigure}

\usepackage{url}

\title{Leveraging Sentence-level Information with Encoder LSTM \\for Semantic Slot Filling}

\author{Gakuto Kurata\\
	IBM Research\\
	{\tt gakuto@jp.ibm.com}
	\And
	Bing Xiang\\
	IBM Watson\\
	{\tt bingxia@us.ibm.com}
	\AND
	Bowen Zhou\\
	IBM Watson\\
	{\tt zhou@us.ibm.com}
	\And
	Mo Yu\\
	IBM Watson\\
	{\tt yum@us.ibm.com}
}

\date{}

\begin{document}

\maketitle

 \begin{abstract}
 Recurrent Neural Network (RNN) and one of its specific architectures, Long Short-Term Memory (LSTM), have been widely used for sequence labeling.
 Explicitly modeling output label dependencies on top of RNN/LSTM is a widely-studied and effective extension.
 We propose another extension to incorporate the global information spanning over the whole input sequence.
 The proposed method, {\em encoder-labeler LSTM}, first encodes the whole input sequence into a fixed length vector with the encoder LSTM, and then uses this encoded vector as the initial state of another LSTM for sequence labeling.
With this method, we can predict the label sequence while taking the whole input sequence information into consideration.
 In the experiments of a slot filling task, which is an essential component of natural language understanding, with using the standard ATIS corpus, we achieved the state-of-the-art $F_{1}$-score of 95.66\%.
 \end{abstract}

\section{Introduction}
\label{sec:introduction}

Natural language understanding (NLU) is an essential component of natural human computer interaction and typically consists of identifying the intent of the users (intent classification) and extracting the associated semantic slots (slot filling)~\cite{de2008spoken}.
We focus on the latter semantic slot filling task in this paper.

Slot filling can be framed as a sequential labeling problem in which the most probable semantic slot labels are estimated for each word of the given word sequence.
%
Slot filling is a traditional task and tremendous efforts have been done, especially since the 1980s when the Defense Advanced Research Program Agency (DARPA) Airline Travel Information System (ATIS) projects started~\cite{price1990evaluation}.
Following the success of deep learning~\cite{hinton2006fast,bengio2009learning}, 
Recurrent Neural Network (RNN)~\cite{elman1990finding,jordan1997serial} and one of its specific architectures, Long Short-Term Memory (LSTM)~\cite{hochreiter1997long}, have been widely used since they can capture temporal dependencies~\cite{yao2013recurrent,yao2014spoken,mesnil2015using}.
The RNN/LSTM-based slot filling has been extended to be combined with explicit modeling of label dependencies~\cite{yao2014recurrent,liurecurrent}.

In this paper, we extend the LSTM-based slot filling to consider sentence-level information.
In the field of machine translation, an encoder-decoder LSTM has been gaining attention~\cite{sutskever2014sequence}, where the encoder LSTM encodes the global information spanning over the whole input sentence in its last hidden state.
Inspired by this idea, we propose an {\em encoder-labeler LSTM} that leverages the encoder LSTM for slot filling.
First, we encode the input sentence into a fixed length vector by the encoder LSTM.
Then, we predict the slot label sequence by the labeler LSTM whose hidden state is initialized with the encoded vector by the encoder LSTM.
With this encoder-labeler LSTM, we can predict the label sequence while taking the sentence-level information into consideration.



The main contributions of this paper are two-folds:
\begin{enumerate}
 \item Proposed an encoder-labeler LSTM to leverage sentence-level information for slot filling.
 \item Achieved the state-of-the-art $F_{1}$-score of 95.66\% in the slot filling task of the standard ATIS corpus.
\end{enumerate}

\section{Proposed Method}
\label{sec:proposed_method}

We first revisit the LSTM for slot filling and enhance this to explicitly model label dependencies.
Then we explain the proposed encoder-labeler LSTM.

\setlength{\subfigcapskip}{-2mm}

\begin{figure*}[t]
  \begin{minipage}[b]{0.275\textwidth}
  	\centering\subfigure[Labeler LSTM(W).]
  		{\includegraphics[width=0.95\columnwidth]{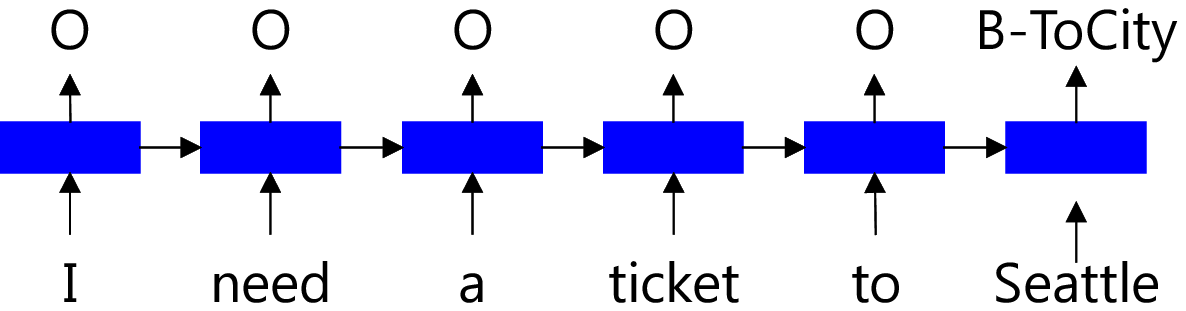}
  		\label{fig:conventional_labeler_lstm}}
  \end{minipage}
  \begin{minipage}[b]{0.275\textwidth}
  	\centering\subfigure[Labeler LSTM(W+L).]
  	{\includegraphics[width=0.95\columnwidth]{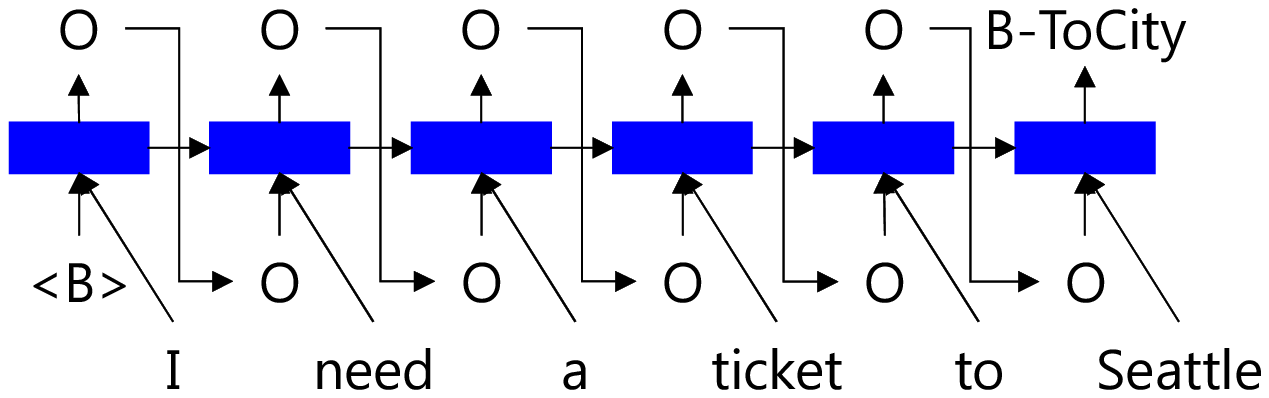}
  		\label{fig:labeler_lstm_w_l}}
  \end{minipage}
  \begin{minipage}[b]{0.45\textwidth}
  	\centering\subfigure[Encoder-decoder LSTM.]
  	{\includegraphics[width=1.0\columnwidth]{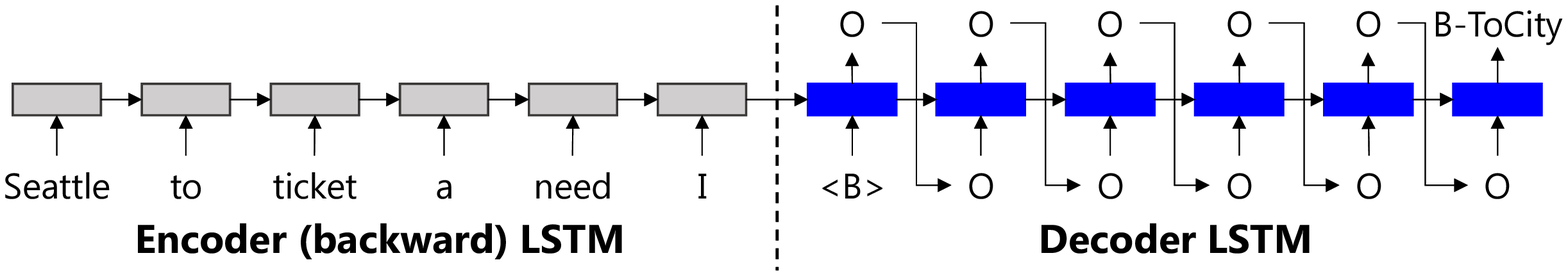}
  		\label{fig:seq_to_seq}}
  \end{minipage}
  
    \begin{minipage}[b]{0.48\textwidth}
    	\centering\subfigure[Encoder-labeler LSTM(W).]
    	{\includegraphics[width=1.0\columnwidth]{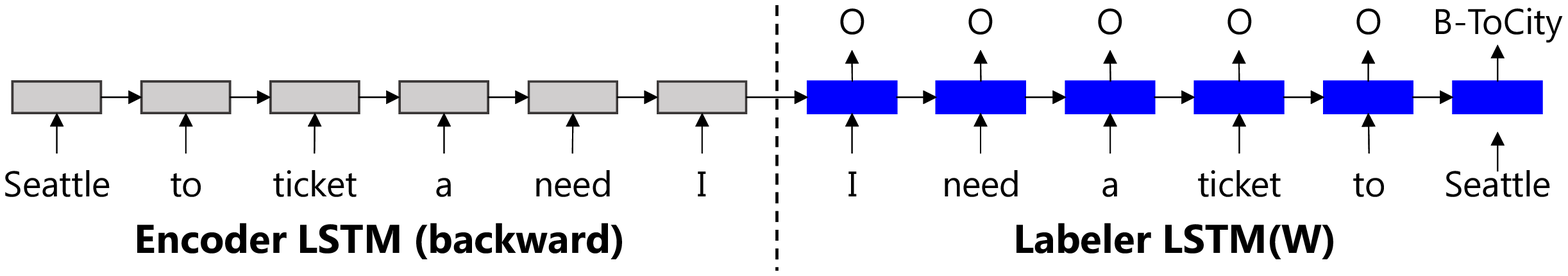}
    		\label{fig:lstm_double_sweep}}
    \end{minipage}
        \begin{minipage}[b]{0.04\textwidth}
        	~
        \end{minipage}
    \begin{minipage}[b]{0.48\textwidth}
    	\centering\subfigure[Encoder-labeler LSTM(W+L).]
    	{\includegraphics[width=1.0\columnwidth]{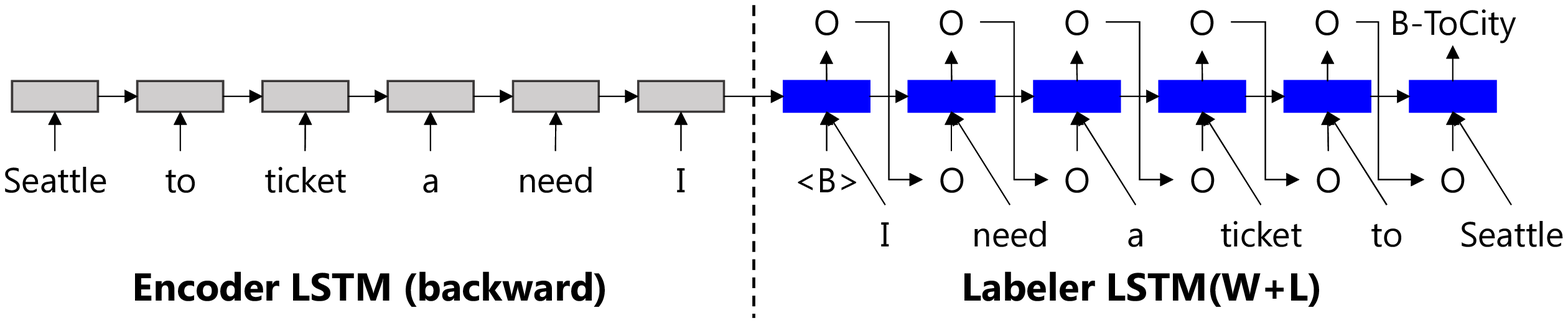}
    		\label{fig:lstm_double_sweep_w_l}}
    \end{minipage}
    \caption{Neural network architectures for slot filling. Input sentence is ``I need a ticket to Seattle''. ``B-ToCity'' is slot label for specific meaning and ``O''is slot label without specific meaning. ``$<$B$>$'' is beginning symbol for slot sequence.}
    \label{fig:nn_architectures}
\end{figure*}

\subsection{LSTM for Slot Filling}
\label{sec:lstm_for_slot_filling}

\figref{fig:conventional_labeler_lstm} shows a typical LSTM for slot filling and we call this as {\em labeler LSTM(W)} where words are fed to the LSTM~\cite{yao2014spoken}.

Slot filling is a sequential labeling task to map a sequence of $T$ words $x_{1}^{T}$ to a sequence of $T$ slot labels $y_{1}^{T}$.
Each word $x_{t}$ is represented with a $V$ dimensional one-hot-vector where $V$ is the vocabulary size and is transferred to $d_{e}$ dimensional continuous space by the word embedding matrix $E \in \mathbb{R}^{d_{e} \times V}$ as $Ex_{t}$.
Instead of simply feeding $Ex_{t}$ into the LSTM, {\em Context Window} is a widely used technique to jointly consider $k$ preceding and succeeding words as $Ex^{t+k}_{t-k} \in \mathbb{R}^{d_{e}(2k+1)}$.
The LSTM has the architecture based on \newcite{jozefowicz2015empirical} that does not have peephole connections and yields the hidden state sequence $h_{1}^{T}$.
For each time step $t$, the posterior probabilities for each slot label are calculated by the softmax layer over the hidden state $h_{t}$.
The word embedding matrix $E$, LSTM parameters, and softmax layer parameters are estimated to minimize the negative log likelihood over the correct label sequences with Back-Propagation Through Time (BPTT)~\cite{williams1990efficient}.

\subsection{Explicit Modeling of Label Dependency}
\label{sec:label_dependency}
A shortcoming of the labeler LSTM(W) is that it does not consider label dependencies.
To explicitly model label dependencies, we introduce a new architecture, {\em labeler LSTM (W+L)}, as shown in \figref{fig:labeler_lstm_w_l}, where the output label of previous time step is fed to the hidden state of current time step, jointly with words, as \newcite{mesnil2015using} and \newcite{liurecurrent} tried with RNN.
For model training, one-hot-vector of ground truth label of previous time step is fed to the hidden state of current time step and for evaluation, left-to-right beam search is used.

\subsection{Encoder-labeler LSTM for Slot Filling}
\label{sec:lstm_encoder}
We propose two types of the encoder-labeler LSTM that uses the labeler LSTM(W) and the labeler LSTM(W+L).
\figref{fig:lstm_double_sweep} shows the {\em encoder-labeler LSTM(W)}.
The encoder LSTM, to the left of the dotted line, reads through the input sentence backward.
Its last hidden state contains the encoded information of the input sentence.
The labeler LSTM(W), to the right of the dotted line, is the same with the labeler LSTM(W) explained in \secref{sec:lstm_for_slot_filling}, {\em except that its hidden state is initialized with the last hidden state of the encoder LSTM.}
The labeler LSTM(W) predicts the slot label conditioned on the encoded information by the encoder LSTM, which means that slot filling is conducted with taking sentence-level information into consideration.
\figref{fig:lstm_double_sweep_w_l} shows the {\em encoder-labeler LSTM(W+L)}, which uses the labeler LSTM(W+L) and predicts the slot label considering sentence-level information and label dependencies jointly.

Model training is basically the same as with the baseline labeler LSTM(W), as shown in \secref{sec:lstm_for_slot_filling}, except that the error in the labeler LSTM is propagated to the encoder LSTM with BPTT.

This encoder-labeler LSTM is motivated by the encoder-decoder LSTM that has been applied to machine translation~\cite{sutskever2014sequence}, grapheme-to-phoneme conversion~\cite{yao2015sequence}, text summarization~\cite{nallapatiabstractive} and so on.
The difference is that the proposed encoder-labeler LSTM accepts the same input sequence twice while the usual encoder-decoder LSTM accepts the input sequence once in the encoder.
Note that the LSTMs for encoding and labeling are different in the encoder-labeler LSTM, 
but the same word embedding matrix is used both for the encoder and labeler since the same input sequence is fed twice.

\subsection{Related Work on Considering Sentence-level Information}
\label{sec:relat-work-cons}
Bi-directional RNN/LSTM have been proposed to capture sentence-level information~\cite{mesnil2015using,zhou2015end,ngoc16:_bi_direc_recur_neural_networ}.
While the bi-directional RNN/LSTM model the preceding and succeeding contexts at a specific word and don't explicitly encode the whole sentence, our proposed encoder-labeler LSTM explicitly encodes whole sentence and predicts slots conditioned on the encoded information.

Another method to consider the sentence-level information for slot filling is the attention-based approach~\cite{simonnet15:_explor_atten_based_recur_neural}.
The attention-based approach is novel in aligning two sequences of different length.
However, in the slot filling task where the input and output sequences have the same length and the input word and the output label has strong relations, the effect of introducing ``soft'' attention might become smaller.
Instead, we directly fed the input word into the labeler part with using context window method as explained in \secref{sec:lstm_encoder}.

\section{Experiments}
\label{sec:experiment}

We report two sets of experiments.
First we use the standard ATIS corpus to confirm the improvement by the proposed encoder-labeler LSTM and compare our results with the published results while discussing the related works.
Then we use a large-scale data set to confirm the effect of the proposed method in a realistic use-case.

\subsection{ATIS Experiment}

\subsubsection{Experimental Setup}
\label{sec:experimental_setup}

\begin{figure}[t]
 \begin{center}
  \includegraphics[width=\columnwidth]{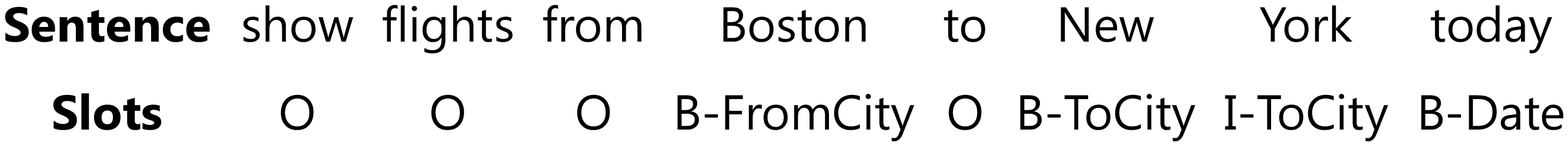}
    \caption{Example of ATIS sentence and annotated slots.}
    \label{fig:atis_example}
 \end{center}
\end{figure}

We used the ATIS corpus, which has been widely used as the benchmark for NLU~\cite{price1990evaluation,dahl1994expanding,wang2006combining,tur2010left}.
\figref{fig:atis_example} shows an example sentence and its semantic slot labels in In-Out-Begin (IOB) representation.
The slot filling task was to predict the slot label sequences from input word sequences.

The performance was measured by the $F_{1}$-score:
$F_{1} = \frac{2 \times Precision \times Recall}{Precision + Recall}$, where precision is the ratio of the correct labels in the system's output and recall is the ratio of the correct labels in the ground truth of the evaluation data~\cite{rijsbergen1979information}.

The ATIS corpus contains the training data of 4,978 sentences and evaluation data of 893 sentences.
The unique number of slot labels is 127 and the vocabulary size is 572.
In the following experiments, we randomly selected 80\% of the original training data to train the model and used the remaining 20\% as the heldout data~\cite{mesnil2015using}.
We reported the $F_{1}$-score on the evaluation data with hyper-parameters that achieved the best $F_{1}$-score on the heldout data.

For training, we randomly initialized parameters in accordance with the normalized initialization~\cite{glorot2010understanding}.
We used {\em ADAM} for learning rate control~\cite{kingma2014adam} and {\em dropout} for generalization with a dropout rate of $0.5$~\cite{srivastava2014dropout,zaremba2014recurrent}.

\subsubsection{Improvement by Encoder-labeler LSTM}
\label{sec:initial_experiments}

We conducted experiments to compare the labeler LSTM(W) (\secref{sec:lstm_for_slot_filling}), the labeler LSTM(W+L) (\secref{sec:label_dependency}), and the encoder-labeler LSTM (\secref{sec:lstm_encoder}).
As for yet another baseline, we tried the encoder-decoder LSTM as shown in \figref{fig:seq_to_seq}\footnote{Length of the output label sequence is equal to that of the input word sequence in a slot filling task. Therefore, ending symbol for slot sequence is not necessary.}.

For all architectures, we set the initial learning rate to $0.001$~\cite{kingma2014adam} and the dimension of word embeddings to $d_{e}=30$.
We changed the number of hidden units in the LSTM, $d_{h} \in \{100, 200, 300\}$\footnote{When using deep architecture later in this section, $d_{h}$ was tuned for each layer.}, and the size of the context window, $k \in \{0,1,2\}$\footnote{In our preliminary experiments with using the labeler LSTM(W), $F_{1}$-scores deteriorated with $k \geq 3$.}.
We used backward encoding for the encoder-decoder LSTM and the encoder-labeler LSTM as suggested in \newcite{sutskever2014sequence}.
For the encoder-decoder LSTM, labeler LSTM(W+L), and encoder-labeler LSTM(W+L), we used the left-to-right beam search decoder~\cite{sutskever2014sequence} with beam sizes of $1$, $2$, $4$, and $8$ for evaluation where the best $F_{1}$-score was reported.
During $100$ training epochs, we reported the $F_{1}$-score on the evaluation data with the epoch when the $F_{1}$-score for the heldout data was maximized.
\tabref{tab:results} shows the results.

The proposed encoder-labeler LSTM(W) and encoder-labeler LSTM(W+L) both outperformed the labeler LSTM(W) and labeler LSTM(W+L), which confirms the novelty of considering sentence-level information with the encoder LSTM by our proposed method.

Contrary to expectations, $F_{1}$-score by the encoder-labeler LSTM(W+L) was not improved from that by the encoder-labeler LSTM(W).
A possible reason for this is the propagation of label prediction errors.
We compared the label prediction accuracy for the words after the first label prediction error in the evaluation sentences and confirmed that the accuracy deteriorated from 84.0\% to 82.6\% by using pthe label dependencies.

For the encoder-labeler LSTM(W) which was better than the encoder-labeler LSTM(W+L), we tried the deep architecture of 2 LSTM layers ({\em Encoder-labeler deep LSTM(W)}).
We also trained the corresponding {\em labeler deep LSTM(W)}.
As in \tabref{tab:results}, we obtained improvement from 94.91\% to 95.47\% by the proposed encoder-labeler deep LSTM(W), which was statistically significant at the 90\% level.

Lastly, $F_{1}$-score by the encoder-decoder LSTM was worse than other methods as shown in the first row of \tabref{tab:results}.
Since the slot label is closely related with the input word, the encoder-decoder LSTM was not an appropriate approach for the slot filling task.

\begin{table}[t]
 \begin{center}
  \begin{tabular*}{\columnwidth}{@{\extracolsep{\fill}}clc}
	\thline
	~&~ & $\bm{ F_{1}}${\bf -score}  \\
    \thline
    (c)& Encoder-decoder LSTM & 80.11 \\
    \hline
  	(a)& Labeler LSTM(W)  & 94.80 \\
   	(d)& {\bf Encoder-labeler LSTM(W)} & {\bf 95.29} \\
  	 \hline
  	(b)& Labeler LSTM(W+L)  & 94.91 \\
	(e)& {\bf Encoder-labeler LSTM(W+L)} & {\bf 95.19} \\
  	\hline
  	~& Labeler Deep LSTM(W) & 94.91  \\
    	~& {\bf Encoder-labeler Deep LSTM(W)} & {\bf 95.47} \\
  	\thline
  \end{tabular*}
  \caption{Experimental results on ATIS slot filling task. Leftmost column corresponds to \figref{fig:nn_architectures}. Lines with bold fonts use proposed encoder-labeler LSTM.~[\%]}
    \label{tab:results}
 \end{center}
\end{table}

\subsubsection{Comparison with Published Results}
\label{sec:comparison_with_related_work}

\tabref{tab:results_previous} summarizes the recently published results on the ATIS slot filling task and compares them with the results from the proposed methods.

Recent research has been focusing on RNN and its extensions.
\newcite{yao2013recurrent} used RNN and outperformed methods that did not use neural networks, such as SVM~\cite{raymond2007generative} and CRF~\cite{deng2012use}.
\newcite{mesnil2015using} tried bi-directional RNN, but reported degradation comparing with their single-directional RNN (94.98\%).
\newcite{yao2014spoken} introduced LSTM and deep LSTM and obtained improvement over RNN.
\newcite{peng2015recurrent} proposed RNN-EM that used an external memory architecture to improve the memory capability of RNN.

Many studies have been also conducted to explicitly model label dependencies.
\newcite{xu2013convolutional} proposed CNN-CRF that explicitly models the dependencies of the output from CNN.
\newcite{mesnil2015using} used hybrid RNN that combined Elman-type and Jordan-type RNNs.
\newcite{liurecurrent} used the output label for the previous word to model label dependencies (RNN-SOP).

\newcite{ngoc16:_bi_direc_recur_neural_networ} recently proposed to use ranking loss function over bi-directional RNNs with achieving 95.47\% (R-biRNN) and reported 95.56\% by ensemble (5$\times$R-biRNN).

By comparing with these methods, the main difference of our proposed encoder-labeler LSTM is the use of encoder LSTM to leverage sentence-level information~\footnote{Since \newcite{simonnet15:_explor_atten_based_recur_neural} did not report the experimental results on ATIS, we could not experimentally compare our result with their attention-based approach. Theoretical comparison is available in \secref{sec:relat-work-cons}.}.

For our encoder-labeler LSTM(W) and encoder-labeler deep LSTM(W), we further conducted hyper-parameter search with a random search strategy~\cite{bergstra2012random}.
We tuned the dimension of word embeddings, $d_{e} \in \{30,50,75\}$, number of hidden states in each layer, $d_{h} \in \{100,150,200,250,300\}$,
size of context window, $k \in \{0,1,2\}$, and initial learning rate sampled from uniform distribution in range $[0.0001,0.01]$.
To the best of our knowledge, the previously published best $F_{1}$-score was 95.56\%\footnote{
There are other published results that achieved better $F_{1}$-scores by using other information on top of word features.
\newcite{vukotic2015time} achieved 96.16\% $F_{1}$-score by using the named entity (NE) database when estimating word embeddings.
\newcite{yao2013recurrent} and \newcite{yao2014spoken} used NE features in addition to word features and obtained improvement with both the RNN and LSTM upto 96.60\% $F_{1}$-score.
\newcite{mesnil2015using} also used NE features and reported $F_{1}$-score of 96.29\% with RNN and 96.46\% with Recurrent CRF.
}~\cite{ngoc16:_bi_direc_recur_neural_networ}.
Our encoder-labeler deep LSTM(W) achieved 95.66\% $F_{1}$-score, outperforming the previously published $F_{1}$-score as shown in \tabref{tab:results_previous}.

Note some of the previous results used whole training data for model training while others used randomly selected 80\% of data for model training and the remaining 20\% for hyper-parameter tuning.
Our results are based on the latter setup.

\begin{table}[t]
 \begin{center}
  \begin{tabular*}{\columnwidth}{@{\extracolsep{\fill}}lc} 
   \thline
   ~ & $\bm{F_{1}}${\bf -score}  \\
   \thline
   RNN~\cite{yao2013recurrent} &  94.11 \\
   CNN-CRF~\cite{xu2013convolutional} & 94.35 \\
   Bi-directional RNN~\cite{mesnil2015using}& 94.73 \\
   LSTM~\cite{yao2014spoken} & 94.85 \\
   RNN-SOP~\cite{liurecurrent} & 94.89 \\
   Hybrid RNN~\cite{mesnil2015using}& 95.06 \\
   Deep LSTM~\cite{yao2014spoken} & 95.08 \\
   RNN-EM~\cite{peng2015recurrent} & 95.25 \\
   R-biRNN~\cite{ngoc16:_bi_direc_recur_neural_networ} & 95.47 \\
   5$\times$R-biRNN~\cite{ngoc16:_bi_direc_recur_neural_networ} & 95.56 \\
   \hline
   {\bf Encoder-labeler LSTM(W)} & {\bf 95.40} \\
   {\bf Encoder-labeler Deep LSTM(W)} & {\bf 95.66} \\
   \thline
  \end{tabular*}
  \caption{Comparison with published results on ATIS slot filling task. $F_{1}$-scores by proposed method are improved from \tabref{tab:results} due to sophisticated hyper-parameters.~[\%]}
  \label{tab:results_previous}
 \end{center}
\end{table}

\subsection{Large-scale Experiment}
\label{sec:large-scale}
We prepared a large-scale data set by merging the MIT Restaurant Corpus and MIT Movie Corpus~\cite{liu2013asgard,liu2013query,spoken13:_mit_restaur_corpus_mit_movie_corpus} with the ATIS corpus.
Since users of the NLU system may provide queries without explicitly specifying their domain, building one NLU model for multiple domains is necessary.
The merged data set contains 30,229 training and 6,810 evaluation sentences.
The unique number of slot labels is 191 and the vocabulary size is 16,049.
With this merged data set, we compared the labeler LSTM(W) and the proposed encoder-labeler LSTM(W) according to the experimental procedure explained in \secref{sec:initial_experiments}.
The labeler LSTM(W) achieved the $F_{1}$-score of 72.80\% and the encoder-labeler LSTM(W) improved it to 74.41\%, which confirmed the effect of the proposed method in large and realistic data set~\footnote{The purpose of this experiment is to confirm the effect of the proposed method. The absolute $F_{1}$-scores can not be compared with the numbers in \newcite{liu2013query} since the capitalization policy and the data size of the training data were different.}.

\section{Conclusion}
\label{sec:conclusion}
We proposed an encoder-labeler LSTM that can conduct slot filling conditioned on the encoded sentence-level information.
We applied this method to the standard ATIS corpus and obtained the state-of-the-art $F_{1}$-score in a slot filling task.
We also tried to explicitly model label dependencies, but it was not beneficial in our experiments, which should be further investigated in our future work.

In this paper, we focused on the slot labeling in this paper.
Previous papers reported that jointly training the models for slot filling and intent classification boosted the accuracy of both tasks~\cite{xu2013convolutional,shi2015contextual,liu2015deep}.
Leveraging our encoder-labeler LSTM approach in joint training should be worth trying.

\section*{Acknowledgments}
We are grateful to Dr. Yuta Tsuboi, Dr. Ryuki Tachibana, and Mr. Nobuyasu Itoh of IBM Research - Tokyo for the fruitful discussion and their comments on this and earlier versions of the paper.
We thank Dr. Ramesh M. Nallapati and Dr. Cicero Nogueira dos Santos of IBM Watson for their valuable suggestions.
We thank the anonymous reviewers for their valuable comments.


\bibliographystyle{emnlp2016}
\bibliography{../../BIB/BIB4INTERNATIONAL}

\begin{thebibliography}{}

\bibitem[\protect\citename{Bengio}2009]{bengio2009learning}
Yoshua Bengio.
\newblock 2009.
\newblock Learning deep architectures for {AI}.
\newblock {\em Foundations and trends{\textregistered} in Machine Learning},
  2(1):1--127.

\bibitem[\protect\citename{Bergstra and Bengio}2012]{bergstra2012random}
James Bergstra and Yoshua Bengio.
\newblock 2012.
\newblock Random search for hyper-parameter optimization.
\newblock {\em The Journal of Machine Learning Research}, 13(1):281--305.

\bibitem[\protect\citename{Dahl \bgroup et al.\egroup }1994]{dahl1994expanding}
Deborah~A Dahl, Madeleine Bates, Michael Brown, William Fisher, Kate
  Hunicke-Smith, David Pallett, Christine Pao, Alexander Rudnicky, and
  Elizabeth Shriberg.
\newblock 1994.
\newblock Expanding the scope of the {ATIS} task: {The ATIS-3} corpus.
\newblock In {\em Proc. HLT}, pages 43--48.

\bibitem[\protect\citename{De~Mori \bgroup et al.\egroup }2008]{de2008spoken}
Renato De~Mori, Fr{\'e}d{\'e}ric Bechet, Dilek Hakkani-Tur, Michael McTear,
  Giuseppe Riccardi, and Gokhan Tur.
\newblock 2008.
\newblock Spoken language understanding.
\newblock {\em IEEE Signal Processing Magazine}, 3(25):50--58.

\bibitem[\protect\citename{Deng \bgroup et al.\egroup }2012]{deng2012use}
Li~Deng, Gokhan Tur, Xiaodong He, and Dilek Hakkani-Tur.
\newblock 2012.
\newblock Use of kernel deep convex networks and end-to-end learning for spoken
  language understanding.
\newblock In {\em Proc. SLT}, pages 210--215.

\bibitem[\protect\citename{Elman}1990]{elman1990finding}
Jeffrey~L Elman.
\newblock 1990.
\newblock Finding structure in time.
\newblock {\em Cognitive science}, 14(2):179--211.

\bibitem[\protect\citename{Glorot and Bengio}2010]{glorot2010understanding}
Xavier Glorot and Yoshua Bengio.
\newblock 2010.
\newblock Understanding the difficulty of training deep feedforward neural
  networks.
\newblock In {\em Proc. AISTATS}, pages 249--256.

\bibitem[\protect\citename{Hinton \bgroup et al.\egroup }2006]{hinton2006fast}
Geoffrey~E Hinton, Simon Osindero, and Yee-Whye Teh.
\newblock 2006.
\newblock A fast learning algorithm for deep belief nets.
\newblock {\em Neural computation}, 18(7):1527--1554.

\bibitem[\protect\citename{Hochreiter and Schmidhuber}1997]{hochreiter1997long}
Sepp Hochreiter and J{\"u}rgen Schmidhuber.
\newblock 1997.
\newblock Long short-term memory.
\newblock {\em Neural computation}, 9(8):1735--1780.

\bibitem[\protect\citename{Jordan}1997]{jordan1997serial}
Michael~I Jordan.
\newblock 1997.
\newblock Serial order: A parallel distributed processing approach.
\newblock {\em Advances in psychology}, 121:471--495.

\bibitem[\protect\citename{Jozefowicz \bgroup et al.\egroup
  }2015]{jozefowicz2015empirical}
Rafal Jozefowicz, Wojciech Zaremba, and Ilya Sutskever.
\newblock 2015.
\newblock An empirical exploration of recurrent network architectures.
\newblock In {\em Proc. ICML}, pages 2342--2350.

\bibitem[\protect\citename{Kingma and Ba}2014]{kingma2014adam}
Diederik Kingma and Jimmy Ba.
\newblock 2014.
\newblock {ADAM}: A method for stochastic optimization.
\newblock {\em arXiv preprint arXiv:1412.6980}.

\bibitem[\protect\citename{Liu and Lane}2015]{liurecurrent}
Bing Liu and Ian Lane.
\newblock 2015.
\newblock Recurrent neural network structured output prediction for spoken
  language understanding.
\newblock In {\em Proc. NIPS Workshop on Machine Learning for Spoken Language
  Understanding and Interactions}.

\bibitem[\protect\citename{Liu \bgroup et al.\egroup }2013a]{liu2013asgard}
Jingjing Liu, Panupong Pasupat, Scott Cyphers, and James Glass.
\newblock 2013a.
\newblock Asgard: A portable architecture for multilingual dialogue systems.
\newblock In {\em Proc. ICASSP}, pages 8386--8390.

\bibitem[\protect\citename{Liu \bgroup et al.\egroup }2013b]{liu2013query}
Jingjing Liu, Panupong Pasupat, Yining Wang, Scott Cyphers, and James Glass.
\newblock 2013b.
\newblock Query understanding enhanced by hierarchical parsing structures.
\newblock In {\em Proc. ASRU}, pages 72--77.

\bibitem[\protect\citename{Liu \bgroup et al.\egroup }2015]{liu2015deep}
Chunxi Liu, Puyang Xu, and Ruhi Sarikaya.
\newblock 2015.
\newblock Deep contextual language understanding in spoken dialogue systems.
\newblock In {\em Proc. INTERSPEECH}, pages 120--124.

\bibitem[\protect\citename{Mesnil \bgroup et al.\egroup }2015]{mesnil2015using}
Gr{\'e}goire Mesnil, Yann Dauphin, Kaisheng Yao, Yoshua Bengio, Li~Deng, Dilek
  Hakkani-Tur, Xiaodong He, Larry Heck, Gokhan Tur, Dong Yu, et~al.
\newblock 2015.
\newblock Using recurrent neural networks for slot filling in spoken language
  understanding.
\newblock {\em IEEE/ACM Transactions on Audio, Speech, and Language
  Processing}, 23(3):530--539.

\bibitem[\protect\citename{Nallapati \bgroup et al.\egroup
  }2016]{nallapatiabstractive}
Ramesh Nallapati, Bowen Zhou, {\c{C}}a~glar Gul{\c{c}}ehre, and Bing Xiang.
\newblock 2016.
\newblock Abstractive text summarization using sequence-to-sequence {RNNs} and
  beyond.
\newblock In {\em Proc. CoNLL}.

\bibitem[\protect\citename{Peng and Yao}2015]{peng2015recurrent}
Baolin Peng and Kaisheng Yao.
\newblock 2015.
\newblock Recurrent neural networks with external memory for language
  understanding.
\newblock {\em arXiv preprint arXiv:1506.00195}.

\bibitem[\protect\citename{Price}1990]{price1990evaluation}
Patti Price.
\newblock 1990.
\newblock Evaluation of spoken language systems: {The ATIS} domain.
\newblock In {\em Proc. {DARPA} Speech and Natural Language Workshop}, pages
  91--95.

\bibitem[\protect\citename{Raymond and Riccardi}2007]{raymond2007generative}
Christian Raymond and Giuseppe Riccardi.
\newblock 2007.
\newblock Generative and discriminative algorithms for spoken language
  understanding.
\newblock In {\em Proc. INTERSPEECH}, pages 1605--1608.

\bibitem[\protect\citename{Shi \bgroup et al.\egroup }2015]{shi2015contextual}
Yangyang Shi, Kaisheng Yao, Hu~Chen, Yi-Cheng Pan, Mei-Yuh Hwang, and Baolin
  Peng.
\newblock 2015.
\newblock Contextual spoken language understanding using recurrent neural
  networks.
\newblock In {\em Proc. ICASSP}, pages 5271--5275.

\bibitem[\protect\citename{Simonnet \bgroup et al.\egroup
  }2015]{simonnet15:_explor_atten_based_recur_neural}
Edwin Simonnet, Camelin Nathalie, Del{\'e}glise Paul, and Est{\`e}ve Yannick.
\newblock 2015.
\newblock Exploring the use of attention-based recurrent neural networks for
  spoken language understanding.
\newblock In {\em Proc. NIPS Workshop on Machine Learning for Spoken Language
  Understanding and Interactions}.

\bibitem[\protect\citename{{Spoken Laungage Systems
  Group}}2013]{spoken13:_mit_restaur_corpus_mit_movie_corpus}
{Spoken Laungage Systems Group}.
\newblock 2013.
\newblock {The MIT Restaurant Corpus} and {The MIT Movie Corpus}.
\newblock {https://groups.csail.mit.edu/sls/downloads/}, {MIT Computer
  Science and Artificial Intelligence Laboratory}.

\bibitem[\protect\citename{Srivastava \bgroup et al.\egroup
  }2014]{srivastava2014dropout}
Nitish Srivastava, Geoffrey Hinton, Alex Krizhevsky, Ilya Sutskever, and Ruslan
  Salakhutdinov.
\newblock 2014.
\newblock Dropout: A simple way to prevent neural networks from overfitting.
\newblock {\em The Journal of Machine Learning Research}, 15(1):1929--1958.

\bibitem[\protect\citename{Sutskever \bgroup et al.\egroup
  }2014]{sutskever2014sequence}
Ilya Sutskever, Oriol Vinyals, and Quoc~VV Le.
\newblock 2014.
\newblock Sequence to sequence learning with neural networks.
\newblock In {\em Proc. NIPS}, pages 3104--3112.

\bibitem[\protect\citename{Tur \bgroup et al.\egroup }2010]{tur2010left}
Gokhan Tur, Dilek Hakkani-Tur, and Larry Heck.
\newblock 2010.
\newblock What is left to be understood in {ATIS}?
\newblock In {\em Proc. SLT}, pages 19--24.

\bibitem[\protect\citename{van Rijsbergen}1979]{rijsbergen1979information}
Cornelis~Joost van Rijsbergen.
\newblock 1979.
\newblock {\em Information Retrieval}.
\newblock Butterworth.

\bibitem[\protect\citename{Vu \bgroup et al.\egroup
  }2016]{ngoc16:_bi_direc_recur_neural_networ}
Ngoc~Thang Vu, Pankaj Gupta, Heike Adel, and Hinrich Sch{\"u}tze.
\newblock 2016.
\newblock Bi-directional recurrent neural network with ranking loss for spoken
  language understanding.
\newblock In {\em Proc. ICASSP}, pages 6060--6064.

\bibitem[\protect\citename{Vukotic \bgroup et al.\egroup
  }2015]{vukotic2015time}
Vedran Vukotic, Christian Raymond, and Guillaume Gravier.
\newblock 2015.
\newblock Is it time to switch to word embedding and recurrent neural networks
  for spoken language understanding?
\newblock In {\em Proc. INTERSPEECH}, pages 130--134.

\bibitem[\protect\citename{Wang \bgroup et al.\egroup }2006]{wang2006combining}
Ye-Yi Wang, Alex Acero, Milind Mahajan, and John Lee.
\newblock 2006.
\newblock Combining statistical and knowledge-based spoken language
  understanding in conditional models.
\newblock In {\em Proc. COLING-ACL}, pages 882--889.

\bibitem[\protect\citename{Williams and Peng}1990]{williams1990efficient}
Ronald~J Williams and Jing Peng.
\newblock 1990.
\newblock An efficient gradient-based algorithm for on-line training of
  recurrent network trajectories.
\newblock {\em Neural Computation}, 2(4):490--501.

\bibitem[\protect\citename{Xu and Sarikaya}2013]{xu2013convolutional}
Puyang Xu and Ruhi Sarikaya.
\newblock 2013.
\newblock Convolutional neural network based triangular {CRF} for joint intent
  detection and slot filling.
\newblock In {\em Proc. ASRU}, pages 78--83.

\bibitem[\protect\citename{Yao and Zweig}2015]{yao2015sequence}
Kaisheng Yao and Geoffrey Zweig.
\newblock 2015.
\newblock Sequence-to-sequence neural net models for grapheme-to-phoneme
  conversion.
\newblock {\em Proc. INTERSPEECH}, pages 3330--3334.

\bibitem[\protect\citename{Yao \bgroup et al.\egroup }2013]{yao2013recurrent}
Kaisheng Yao, Geoffrey Zweig, Mei-Yuh Hwang, Yangyang Shi, and Dong Yu.
\newblock 2013.
\newblock Recurrent neural networks for language understanding.
\newblock In {\em Proc. INTERSPEECH}, pages 2524--2528.

\bibitem[\protect\citename{Yao \bgroup et al.\egroup }2014a]{yao2014spoken}
Kaisheng Yao, Baolin Peng, Yu~Zhang, Dong Yu, Geoffrey Zweig, and Yangyang Shi.
\newblock 2014a.
\newblock Spoken language understanding using long short-term memory neural
  networks.
\newblock In {\em Proc. SLT}, pages 189--194.

\bibitem[\protect\citename{Yao \bgroup et al.\egroup }2014b]{yao2014recurrent}
Kaisheng Yao, Baolin Peng, Geoffrey Zweig, Dong Yu, Xiaolong Li, and Feng Gao.
\newblock 2014b.
\newblock Recurrent conditional random field for language understanding.
\newblock In {\em Proc. ICASSP}, pages 4077--4081.

\bibitem[\protect\citename{Zaremba \bgroup et al.\egroup
  }2014]{zaremba2014recurrent}
Wojciech Zaremba, Ilya Sutskever, and Oriol Vinyals.
\newblock 2014.
\newblock Recurrent neural network regularization.
\newblock {\em arXiv preprint arXiv:1409.2329}.

\bibitem[\protect\citename{Zhou and Xu}2015]{zhou2015end}
Jie Zhou and Wei Xu.
\newblock 2015.
\newblock End-to-end learning of semantic role labeling using recurrent neural
  networks.
\newblock In {\em Proc. ACL}, pages 1127--1137.

\end{thebibliography}

\end{document}